\documentclass[10pt,twocolumn,letterpaper]{article}

\usepackage{arxiv}              %

\usepackage{times}
\usepackage{epsfig}
\usepackage{graphicx}
\usepackage{amsmath}
\usepackage{amssymb}
\usepackage{algpseudocode,algorithm,algorithmicx}

\newcommand*\Let[2]{\State #1 $\gets$ #2}
\algrenewcommand\algorithmicrequire{\textbf{Inputs:}}
\algrenewcommand\algorithmicensure{\textbf{Postcondition:}}
\def\sf{Segment-Fusion\xspace}

\usepackage{booktabs}

\usepackage[pagebackref=true,breaklinks=true,colorlinks,bookmarks=false]{hyperref}
\usepackage[capitalize]{cleveref}
\crefname{section}{Sec.}{Secs.}
\Crefname{section}{Section}{Sections}
\Crefname{table}{Table}{Tables}
\crefname{table}{Tab.}{Tabs.}

\newcommand{\tablenegspace}{\vspace{-4mm}} %

\def\confName{}
\def\rot{\rotatebox}

\graphicspath{{./images/}}

\begin{document}

\title{Robust 3D Scene Segmentation through Hierarchical and Learnable Part-Fusion}

\author{Anirud Thyagharajan, Benjamin Ummenhofer, Prashant Laddha, Om J Omer, Sreenivas Subramoney\\
        Intel Labs}
\affiliation{Intel Labs}

\maketitle

\begin{abstract}
   3D semantic segmentation is a fundamental building block for several scene
   understanding applications such as autonomous driving, robotics and AR/VR.
   Several state-of-the-art semantic segmentation
   models suffer from the part-misclassification problem, wherein parts of the same object are labelled incorrectly.
   Previous methods have utilized hierarchical,
   iterative methods to fuse semantic and instance information, but they lack learnability in context fusion,
   and are computationally complex and heuristic driven.
   This paper presents \sf, a novel attention-based method for hierarchical fusion of semantic and instance information to address the part misclassifications.
   The presented method includes a graph segmentation algorithm for grouping points into segments that pools point-wise features
   into segment-wise features, a learnable attention-based network to fuse these segments based on their semantic and instance features,
   and followed by a simple yet effective connected component labelling algorithm to convert segment features to instance labels.
   \sf  can be flexibly employed with any network architecture for semantic/instance segmentation.
   It improves the qualitative and quantitative performance of several semantic segmentation backbones
   by upto $5\%$ when evaluated on the ScanNet and S3DIS datasets. %

\end{abstract}

\section{Introduction}
The growing resolution and availability of 3D visual sensors in recent years (e.g. Kinect, RealSense, Xtion) have enabled high fidelity representation of real world scenes using 3D data, allowing machines to understand 3D scenes with higher accuracy.
One of the most important tasks in scene understanding includes 3D semantic segmentation, which aims to recognize the object class that each point in the scene belongs to.
3D semantic segmentation is fundamental to various applications, including but not limited to autonomous driving, robotic navigation, localization and mapping, and scene understanding ~\cite{siam2018comparative,valada2017adapnet,yu2018ds,gupta2015indoor}.

\begin{figure}[t]
  \centering
  \includegraphics[width=1.0\columnwidth]{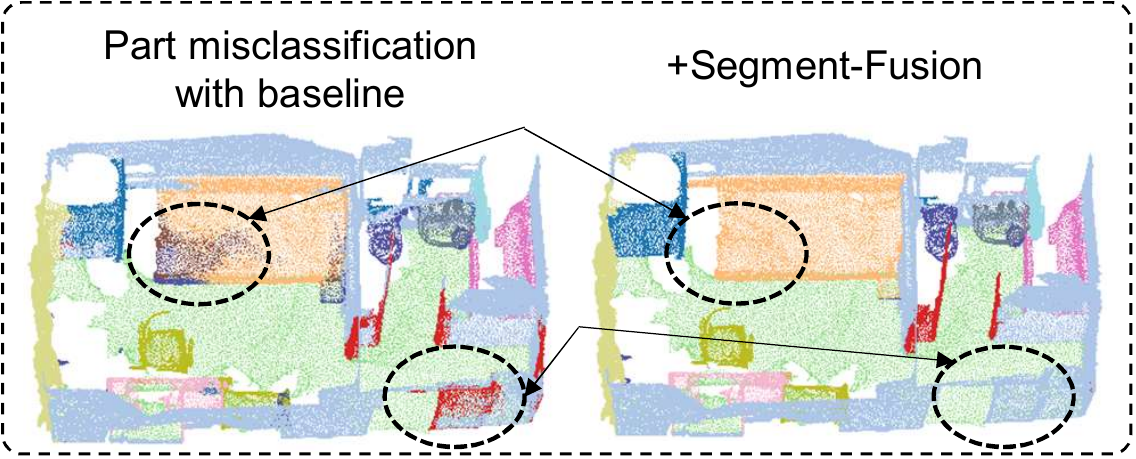}
  \caption{Semantic labels with baseline \cite{choy20194d} and our \sf approach: The highlighted regions show a partly mislabeled object, a common problem for semantic segmentation methods, and the improved result after applying \sf on a sample point cloud
  from the ScanNet validation set \cite{dai2017scannet}.}
  \label{fig:sf_exemplar}
\end{figure}

Recent advancements in 3D scene understanding have been greatly influenced by deep neural networks achieving state-of-the-art results  \cite{wang2017cnn,nekrasov2021mix3d,hu2021bidirectional,hu2021vmnet,kundu2020virtual,gong2021omni,hu2020jsenet,graham20183d,choy20194d,wu2019pointconv}
on multiple benchmarks and wide range of datasets \cite{dai2017scannet,armeni2017joint}.
However, many of these methods are often plagued by the part-misclassification problem, where parts of the same object are labelled incorrectly as shown in Figure~\ref{fig:sf_exemplar}.
This problem can be better addressed if object instance boundaries can be correctly estimated, which includes annotating points with object instance identifiers.
This helps to group all points associated with an object instance and utilize consensus to rectify semantic mispredictions.

Jointly solving semantic and instance segmentation has been explored by several previous works \cite{han2020occuseg,wang2019associatively,jiang2020pointgroup,chen2021hierarchical,wang2018sgpn,pham2019jsis3d,hou20193d,liang2021instance,engelmann20203d}.
Many of these methods \cite{wang2019associatively,jiang2020pointgroup,wang2018sgpn,hou20193d,pham2019jsis3d}
focus on employing models for instance and semantic segmentation, before fusing pointwise features.
However, they require higher compute resources as they fuse at a fine-grained level and do not exploit local continuities in scene structure. %
Methods such as \cite{han2020occuseg,chen2021hierarchical,liang2021instance,engelmann20203d} include hierarchical pooling of features over a spatial region%
addressing the part misclassification problem by fusing these regions correctly, with an assumption that these regions do not stretch over object boundaries.
Such hierarchical methods are more efficient since they work with regions and not directly with points.
\begin{figure*}[t]
  \centering
  \includegraphics[width=0.85\textwidth]{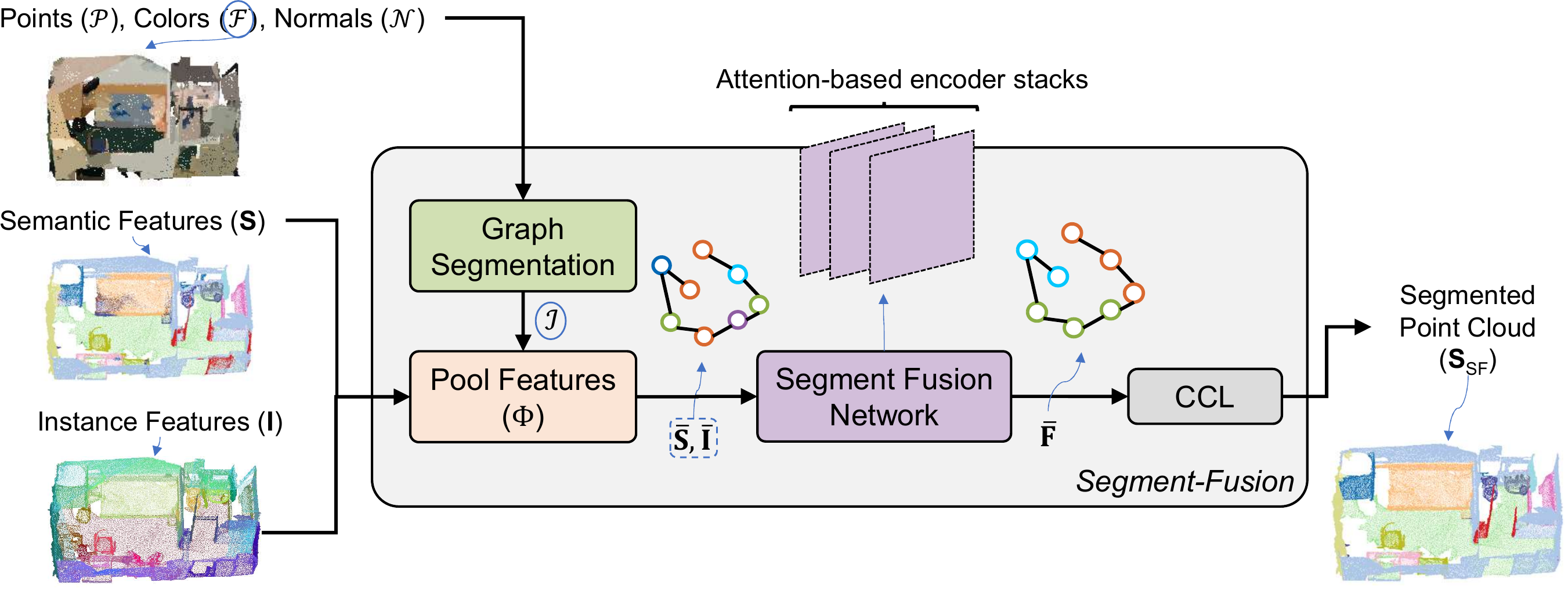}
  \caption{%
  Overview of the \sf approach. %
  The Graph Segmentation module (GS) groups points into segments and enables pooling pointwise semantic and instance features $\mathbf{S}, \mathbf{I}$ into segment-wise features $\bar{\mathbf{S}},\bar{\mathbf{I}}$. %
  The Segment Fusion Network maps these features to a common feature space $\bar{\mathbf{F}}$ which is then used in the connected component labelling algorithm (CCL) to create the final set of segments.%
  }
  \label{fig:seg_fusion_overview}
\end{figure*}

While such methods are more efficient, they employ iterative clustering and aggregation
techniques that are known to be heuristic-driven, non-learnable and specific to the backbone models used for extracting semantic and instance features.
They also use complex post-processing \cite{derpanis2005mean,schubert2017dbscan} to transform instance features to labels.
Thus, there is a need for hierarchical, learnable methods for fusing such regions with simple post-processing.

Our prime objective is to improve semantic segmentation performance of a generic semantic segmentation model by using a hierarchical fusion of semantic and instance information.
We use a graph segmentation algorithm to group points into regions and pool region-wise  semantic and instance features but downstream, we propose a learnable clustering algorithm.
It is desirable that the method should be learnable and agnostic to dataset specific heuristics.
\begin{itemize}
    \item We propose a graph segmentation algorithm optimized for semantic segmentation across datasets, which is used to compose semantic and instance features at a coarser level (which we term as \textit{segments}).
    \item We propose a learnable attention based network, Segment Fusion, to hierarchically fuse these segments based on the similarity of their semantic and instance features, thus understanding the appropriate granularity of context (local to global: points to instance to scene). 
    \item Our approach offers the advantages of adapting to inputs from other datasets and other backbones, as well as offering the possibility to perform end-to-end training and maintain efficiency (while working on a hierarchically smaller representation, and not working on individual points).
   These proposals help ameliorate the problem of part-misclassification by improving the performance (mIoU) of
   multiple semantic backbones on datasets like ScanNet V2 \cite{dai2017scannet} and S3DIS \cite{armeni2017joint} upto $5$\%.
   We also compare the impact of our proposed method with the iterative clustering method proposed by Occuseg \cite{han2020occuseg}
   and report $1$-$2$\% mIoU improvement in semantic segmentation.
\end{itemize}

\section{Related Work}
\label{sec:related_work}

\textbf{3D Semantic Segmentation. } Semantic segmentation techniques can be broadly classified into 2D projection based and 3D methods.
Methods that use 3D processing are known to provide superior accuracy because they can naturally overcome occlusion
or scale ambiguity \cite{han2020occuseg}. Several 3D methods have been proposed for semantic segmentation, which
can be further categorized into point-based \cite{qi2017pointnet,qi2017pointnet++} and volumetric methods \cite{wang2017cnn,graham20183d,choy20194d,hu2021vmnet,hu2021bidirectional,hu2020jsenet}. Methods from the PointNet family \cite{qi2017pointnet,qi2017pointnet++}
work on unstructured
point clouds directly, employing pointwise networks to extract features and using ball queries and hierarchical grouping to encode spatial locality.
\par
\begin{figure}[t]
  \centering
  \includegraphics[width=0.35\textwidth]{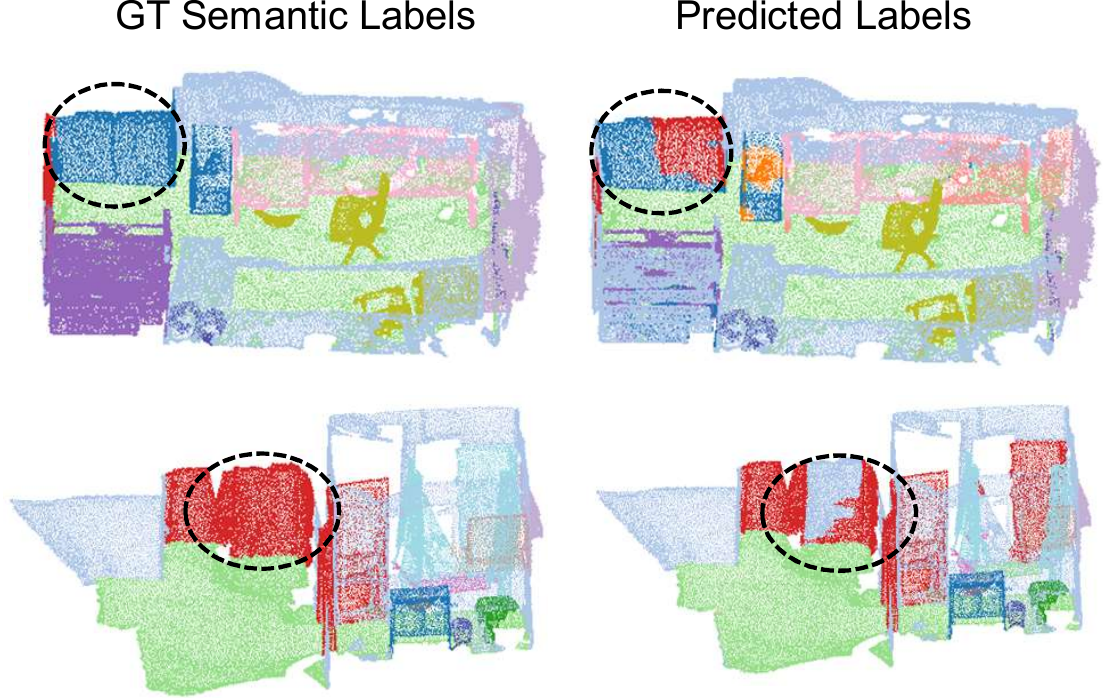}
  \caption{Illustrations of the part misclassification problem while using SCN \cite{graham20183d} (first row) and MinkowskiNet \cite{choy20194d} (second row) as backbones for predicting the semantic classes of the ScanNet V2 dataset \cite{dai2017scannet}.}
  \label{fig:part_misclassification}
\end{figure}

On the other hand, volumetric methods voxelize point clouds into regular grids and process them with regular structured computations.
SparseConvNet \cite{graham20183d} proposed submanifold sparse convolutions to process only the active voxels in a scene, while MinkowskiNet \cite{choy20194d}
generalized this concept with the help of hybrid kernels to express different types of local structure.
Apart from these classes of methods, hybrid methods that utilize deformable/parameteric convolutions \cite{thomas2019kpconv} on points have been proposed, in addition
to employing graph convolutions \cite{hu2020class, wang2019graph, zhang2019dual} on points.
All the above models apply voxel/point-wise cross entropy loss to optimize the model weights. Eventhough they employ network
topologies such as U-Nets that are intended to comprehend scale and local-to-global structure, they exhibit part-misclassification issues
as observed in Figure~\ref{fig:part_misclassification}. Therefore, the model needs additional information about which instance of the object each point
is a part of.

\textbf{Joint Semantic-Instance Segmentation. } Instance segmentation techniques aim to predict features for every point that
indicate the object with which it is associated. Several methods have been proposed to perform semantic and instance
segmentation jointly, since there is shared context between two similar tasks. Methods such as ASIS \cite{wang2019associatively}, SGPN \cite{wang2018sgpn} and JSIS3D \cite{pham2019jsis3d}
exercise models to fuse semantic and instance features pointwise, and subsequently compute the instance labels through
clustering algorithms; ASIS \cite{wang2019associatively} uses mean-shift clustering, SGPN \cite{wang2018sgpn} uses NMS to filter proposals and JSIS3D \cite{pham2019jsis3d} uses a CRF to cluster instance
embeddings into labels. However, these methods are heuristic driven (eg. mean-shift clustering \cite{derpanis2005mean} depends upon the window size)
and are computationally complex (since they work with the pointwise features).
\par
Works such as Occuseg, 3D-MPA and HAIS \cite{engelmann20203d, chen2021hierarchical} simplify this complexity by grouping features
at a higher abstractive level of surfaces (than points), thus proposing hierarchically fusing these features to form instances.
Occuseg \cite{han2020occuseg} computes instance and semantic features per supervoxel, where supervoxels are composed using a %
graph segmentation scheme \cite{felzenszwalb2004efficient}, which are fed to an iterative clustering algorithm to merge supervoxels
into instances. 3D-MPA \cite{engelmann20203d} samples keypoints through a deep voting scheme, uses a graph convolutional network to refine the keypoint
features, which are fed to DBSCAN \cite{schubert2017dbscan} to obtain instance labels. HAIS \cite{chen2021hierarchical} proposed a hierarchical method to generate point sets
(based on the semantic features) and employ an iterative set aggregation algorithm to aggregate these point sets into instances.
SST-Net \cite{liang2021instance} propose a tree classifier to hierarchically build a subtree of superpoint proposals.
\par
While these methods may be effective for certain datasets, they propose iterative algorithms, which are not learnable. This makes it
difficult for the algorithm to generalize for multiple sets of input features and/or datasets. Eventhough 3D-MPA uses a graph convolutional
layer to refine the higher-level features, (a) they work with a fixed number of proposals in a scene (due to the deep voting scheme) which
does not help scale with scenes, and (b) they eventually use a heuristic driven classifier (DBSCAN \cite{schubert2017dbscan}) after the refinement.
\par

\begin{figure}[h]
  \centering
  \includegraphics[width=1.0\columnwidth]{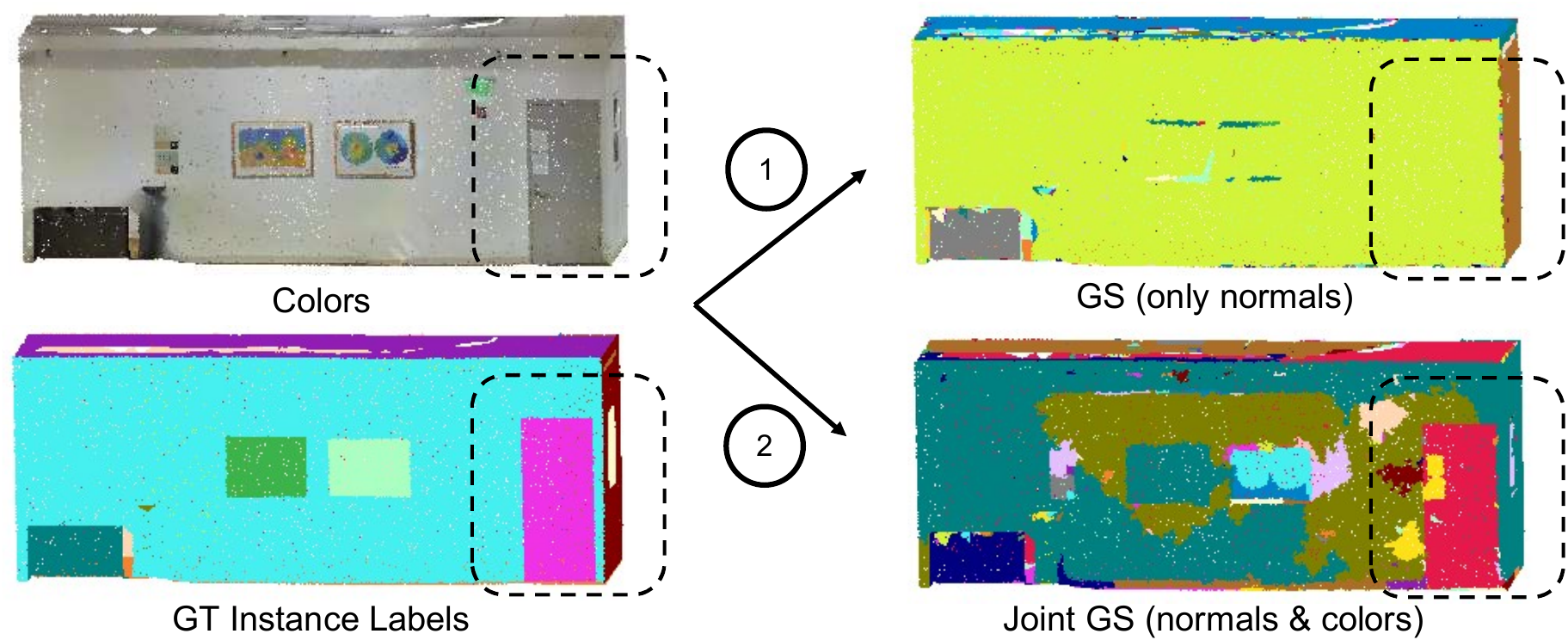}
  \caption{Impact of proposed J-GS method on an example S3DIS point cloud. As observed in \textcircled{1} \cite{felzenszwalb2004efficient}, graph segmentation
           purely based on difference in normals fails to recognise the boundary between the door and the wall, while
           augmenting it with color information helps in discerning this (J-GS \textcircled{2}).}
  \label{fig:s3dis_joint_seg}
\end{figure}

\textbf{Graph Segmentation for hierarchical processing. }As discussed above, graph segmentation algorithms \cite{felzenszwalb2004efficient} have been used by previous methods to group points into larger regions (based on similarity of normals) work at the abstraction level of these regions.
It is observed that representations in higher-level abstractions such as these segments offer geometric continuity,
particularly since they provide guarantees that the point normals in a segment vary only within a bound.
Graph segmentation solely based on normals is effective but in some datasets normals across object boundaries are not distinctive enough, as seen in Figure~\ref{fig:s3dis_joint_seg}.
This problem cannot be solved by lowering threshold for discriminating normals, because it leads to a highly over-segmented
graph and defeats the purpose of hierarchical processing. Previous work \cite{strom2010graph} aims to solve this problem by performing segmentation using dissimilarity in normals and colors,
but does not generalize for other kinds of features, or propose methods for selecting the thresholds for normals/colors.
\par
Ideally, a semantic-instance fusion algorithm should have the following desirable properties:
\begin{itemize}
  \item \textbf{Hierarchical. } To be efficient, the system should abstract features from the points level to a higher-level spatial representation (\textit{segments}: eg. surfaces, supervoxels)
    and process at this level.
  \item \textbf{Learnable. } The task of fusion of features to spatially merge segments into instances should be done by a
    learnable model to accommodate transferring knowledge to different datasets and different semantic and instance backbones.
  \item \textbf{Agnostic to semantic and instance backbones. } The fusion method should be compatible with multiple methods for extracting semantic and instance features.
  \item \textbf{Simple post-processing. } As far as possible, the task of converting instance features to labels should be
    done simplistically to avoid imparting any heuristic driven processing (and absorb learning to the learnable component).
\end{itemize}

\section{Methods}
We present an overview of our method in Figure~\ref{fig:seg_fusion_overview}, attempting to incorporate the desirable properties discussed in Section~\ref{sec:related_work}.
The proposed method assumes a (i) semantic model that emits per-point semantic features $\mathbf{S}$ and (ii) an
instance model that emits per-point instance features $\mathbf{I}$. Our method applies equivocally to those cases where
these models could be sharing a common backbone network too. For instance models like Occuseg \cite{han2020occuseg} that emit
multiple types of embeddings (eg. instance embeddings, centroid and occupancy estimates), $\mathbf{I}$ can be thought of as a
concatenation of such features pointwise.
In general, we denote point-wise features by $\mathbf{X}$ and segment-wise features by $\overline{\mathbf{X}}$.

\subsection{Joint Graph Segmentation (J-GS)}
\label{sec:gsv_method}
We propose a hierarchical strategy, segmenting points into 3D surfaces (segments or super-voxels) using efficient
graph segmentation algorithms such as ~\cite{felzenszwalb2004efficient, strom2010graph, klasing2008clustering}.
The input graph is defined by vertices (points $\mathcal{P}$) joined by edges (neighbors $E$), where the goal of the segmentation
is to find a mapping $\mathcal{J}$ that assigns the same segment ID to a set of grouped points.
The algorithm also records the connections ($\mathcal{E}$) between the segments, reresented by an adjacency matrix $\mathbf{A}$.
As a result of our proposed method, we observe that we obtain over-segmented classification boundaries which largely do not violate object boundaries,
such that points belonging to different objects end up in different segments (Figure~\ref{fig:s3dis_joint_seg}).

To cover a gamut of datasets, we generalize our graph segmentation algorithm to work across two options:
\begin{itemize}
  \item \textbf{Point clouds without mesh information. } Datasets like ScanNet \cite{dai2017scannet} have mesh information
    available, which allows us to compute per vertex normals $\mathcal{N}$ from the adjacent polygonal faces. However, for datasets like
    S3DIS \cite{armeni2017joint} which do not have such connectivity information, we propose obtaining normals using plane-fitting techniques,
    as obtaining mesh information would incur additional expensive pre-processing.

  \item \textbf{Indiscriminative Normals. }
    We propose to generalize this method across datasets over previous work \cite{strom2010graph} by (a) allowing to use arbitrary pointwise features $\mathcal{F}$ (not specifically colors) in addition to normals, 
    and then performing a joint segmentation over two spaces ($\mathcal{F}$ and $\mathcal{N}$), as shown in Algorithm~\ref{alg:graph_seg}.
    An example using colors as $\mathcal{F}$ for the S3DIS dataset is shown in Figure~\ref{fig:s3dis_joint_seg}.
    (b) we generalize sorting orders ($s: \{norm, feats\}$, choosing to sort edges along similarity in $\mathcal{N}$ or $\mathcal{F}$), 
    and (c) we employ a voting-based methodology (see Algorithm~\ref{alg:gsv}) to decide the optimal similarity threshold for a generic dataset for the semantic segmentation task.
\end{itemize}

In Algorithm~\ref{alg:graph_seg}, \texttt{computeEdgeWeights} computes dissimilarity scores along $\mathcal{N}$ (cosine product) and $\mathcal{F}$ ($\ell_1$/$\ell_2$ norms).
\texttt{updateThresh} is similar to previous work \cite{strom2010graph} where the threshold slightly increases to accommodate a growing segment,
where it approaches the largest intra-segment weight.

\begin{algorithm}
  \footnotesize
  \caption{Proposed Graph Segmentation
    \label{alg:graph_seg}}
  \begin{algorithmic}[1]
    \Require{Normals $\mathcal{N}$, Features $\mathcal{F}$, Polygonal edges $E$, feature and normal similarity thresholds ($f_{th}, n_{th}$), sort-flag ($s$)}
    \Statex
    \Function{segment\_graph}{$\mathcal{N}, \mathcal{F}, E, n_{th}, f_{th}, s$}
    \Let{$E_{wn}, E_{wf}$}{computeEdgeWeights($\mathcal{N}, \mathcal{F}$)}
    \Let{$E$}{sortEdgesByNormalsOrFeats($s$)}
    \Let{$th_{f}, th_{n}$}{$f_{th}, n_{th}$}\Comment{initialize threshold vectors}
    \Let{$\mathcal{J}$}{initUnionFind($E$)}
    \For{$e_{ij} \in E$} \Comment{edge joining ith and jth points}
      \Let{$S_i$}{$\mathcal{J}$.find(i)}
      \Let{$S_j$}{$\mathcal{J}$.find(j)}
      \Let{$e_{wn}, e_{wf}$}{$E_{wn}[i, j], E_{wf}[i, j]$} \Comment{edge weights}
      \If{($e_{wn} < th_n[i], th_n[j]$) and ($e_{wf} < th_f[i], th_f[j]$)}
        \Let{$newID$}{$\mathcal{J}$.join($i, j$)}
        \State updateThresh($e_{wn}, e_{wf}, th_n, th_f, newID$)
      \EndIf
    \EndFor
    \State \Return{$\mathcal{J}$}
    \EndFunction
  \end{algorithmic}
\end{algorithm}

\begin{algorithm}
  \footnotesize
  \begin{algorithmic}
    \caption{Graph Segmentation based Voting algorithm%
      \label{alg:gsv}}
      \Require{List of J-GS Params $g_j = (n_{th}, f_{th}, s)$, predicted semantic labels $\mathbf{S}$, groundtruth labels $\mathbf{G}$}
      \Statex
      \Function{J-GSV}{$GP, \mathbf{S}$}
        \State init scores
        \For{$g_j$ in $GP$}
          \Let{$\mathcal{J}_j$}{SEGMENT\_GRAPH($g_j$)}
          \Let{$\mathbf{S}_{j, maj}$}{majorityVoting($\mathcal{J}_j, \mathbf{S}$)}
          \Let{scores[j]}{getIoU($\mathbf{S}_{j, maj}, G$)}
        \EndFor
      \State \Return{argmin(scores)}
      \EndFunction
  \end{algorithmic}
\end{algorithm}

\begin{figure}[h]
  \centering
  \includegraphics[width=1.0\columnwidth]{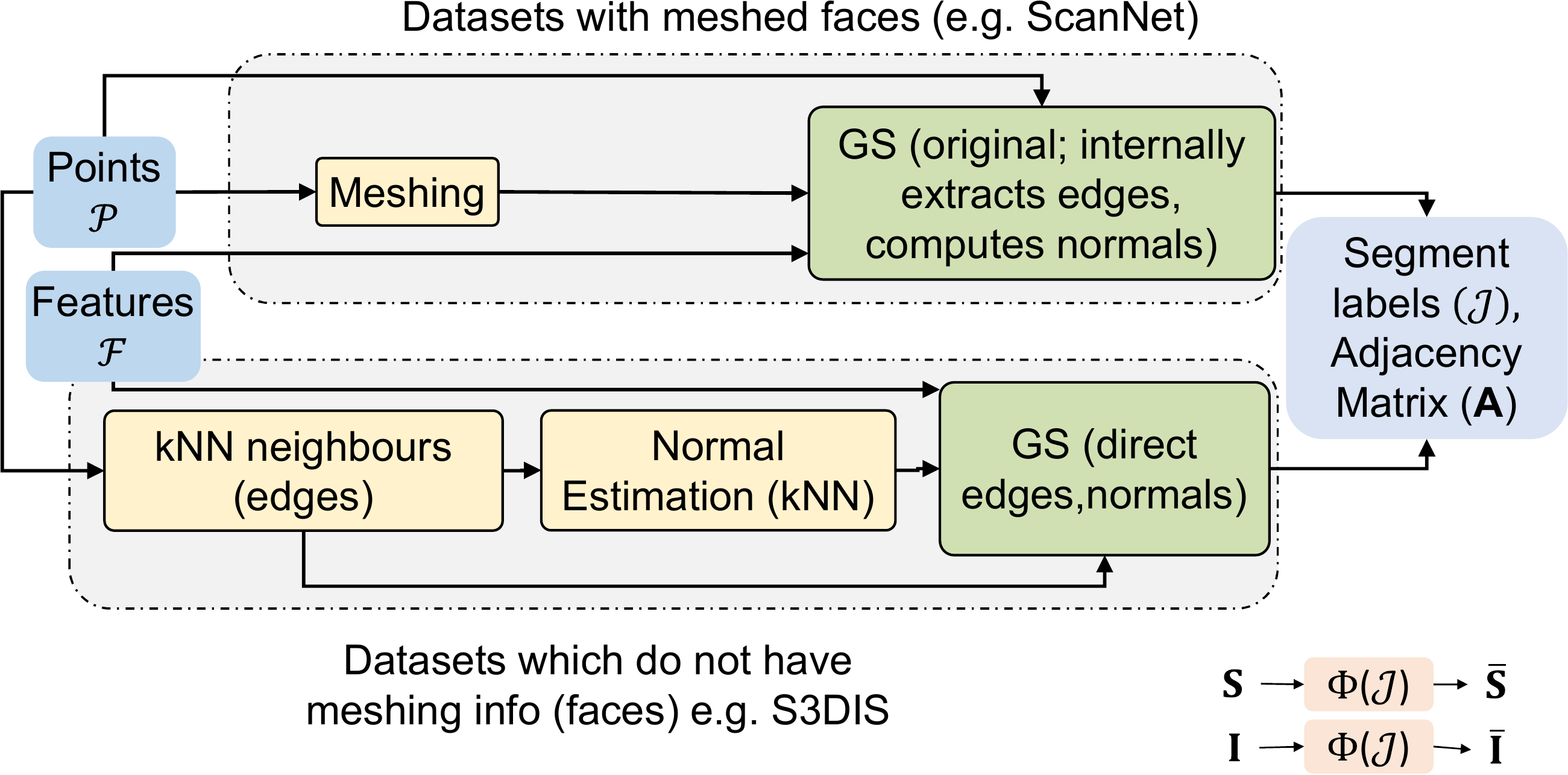}
  \caption{Proposed flow for generalizing  graph segmentation across a variety of point cloud datasets.}
  \label{fig:pcl_non_mesh_flow}
\end{figure}

To select the right variant for given dataset, we enable quantitative comparison across the above options by proposing
the J-GSV score (Joint Graph Segmentation-based Voting) (Algorithm~\ref{alg:gsv}).
Across point clouds from the training set, for the $j$-th variant, we obtain a list of sets of points $\mathcal{J}_j$, with
the $i$-th set $\mathcal{J}_{j, i}$ containing points pertaining to the same segment. We compute the semantic predictions
for the $j$-th variant by pooling the per-point semantic predictions $\mathbf{S}$ for each segment, and broadcasting
the majority label to all points in the segment. The mean Intersection-over-Union (mIoU) for the modified semantic predictions $\mathbf{S}_{j, maj}$
is the score associated with the $j$-th variant. For the dataset, we select the variant with the highest score and denote by $\mathcal{J^{*}}$.
Consequently, we employ a differentiable function $\phi$, to transform
pointwise features to segment-wise features ($\overline{\mathbf{X}} = \phi(\mathbf{X}, \mathcal{J}^{*})$). To simplify, our implementation of $\phi(.)$ is the mean function,
average-pooling across pointwise semantic and instance features ($\mathbf{S},\mathbf{I}$) to obtain segment-wise features ($\bar{\mathbf{S}},\bar{\mathbf{I}}$), similar to \cite{han2020occuseg}.
For pointwise features $\mathbf{X}$ we get the feature for a segment as 
\begin{equation}
  \overline{\mathbf{X}_i} = \frac{1}{|\mathcal{J}_i|}\sum_{j \in \mathcal{J}_i}\mathbf{X}_j.
\end{equation}

Thus, the proposed joint graph segmentation hierarchically composes features of a scene at a coarser level (than at the level of points) to enable efficient fusion decisions of these segments.

\subsection{\sf Network}
\label{sec:nw_details}
Having formed segments out of the underlying points, the next goal is to understand how to form objects out of these
segments. To this effect, we train a network to jointly associate the instance-level and semantic-level information
to essentially form decisions if a pair of segments belong to the same object (\textit{fusable}) or not (\textit{separable}).
\par
We project the instance and semantic features per segment $\{\bar{\mathbf{S}}, \bar{\mathbf{I}}\}$ to a transformed set of features $\bar{\mathbf{F}}$ in the
joint semantic-instance space using the \sf network.%
We compose the \sf network using a stack of attention encoder blocks (taking cues from a Transformer ~\cite{vaswani2017attention} block).
Attention blocks are known to capture larger contextual information among different segments by extracting useful intermediate representations using scaled-dot-product attention.
Additionally, in each block, we propose multiplying the attention-matrix element-wise with the adjacency matrix $\mathbf{A}$ describing the connections
between the segments.
This helps to constrain the interactions only between the segment pairs that are spatially connected.
A more detailed description of the network architecture is provided in the supplementary material.
\par
To supervise this learning process,
we propose two sets of losses - (i) Instance loss and (ii) Segment Loss.
Thus, the overall loss function is:

\begin{equation}
  \mathcal{L}_{SF} = \mathcal{L}_\text{instance} + \mathcal{L}_\text{segment}
\end{equation}

\subsubsection{SF-Instance Loss}
We propose attraction and repulsion instance losses at the segment level. These losses ensure that segments of the
same instance are clustered together (Equation \eqref{eqn:ins_attract}), whereas the centroids of the instance features are repelled from each other (Equation \eqref{eqn:ins_repel}).
While this appears similar to the instance losses employed at the per-point level \cite{han2020occuseg,wang2019associatively} in principle, but in our case, the losses are applied at
the segment-level features. Additionally, we parameterise the thresholds for repulsion and attraction to be $\Delta_D$ and $\Delta_V$ respectively,
essentially to unify the thresholds used in the SF-Segment loss and CCL algorithm.

\begin{equation}
  \mathcal{L}_\text{instance} = \mathcal{L}_\text{attract} + \mathcal{L}_\text{repel} + \mathcal{L}_\text{reg}
\end{equation}

\begin{equation}
  \mathcal{L}_\text{attract} = \frac{1}{K} \sum_{i=1}^{K} \frac{1}{N_i} \sum_{j=1}^{N_i} {H(d(\bar{\mathbf{F}_j}, \mu_i) - \Delta_V)}^2
  \label{eqn:ins_attract}
\end{equation}

\begin{equation}
  \mathcal{L}_\text{repel} = \frac{1}{K (K - 1)} \sum_{i=1}^{K} \sum_{j=1; i \neq j}^{K} {H(\Delta_D - d(\mu_i, \mu_j))}^2
  \label{eqn:ins_repel}
\end{equation}

\begin{equation}
  \mathcal{L}_\text{reg} = \frac{1}{K} \sum_{i=1}^{K} ||\mu_i||_{1}
\end{equation}
where $K$ denotes the number of ground truth instances in the scene; $||.||$ is the $\ell_1$ norm; $\mu_i$ is the average
of the segment features across the segments belonging to the ith instance;
$d(f_i, f_j)$ indicates feature distance using a suitable norm ($\ell_1$/$\ell_2$), and $H(..)$ is the hinge loss;
we set the values of the thresholds $\Delta_V$ and $\Delta_D$ to be 0.10 and 1 respectively.

\subsubsection{SF-Segment Loss}
The SF-instance losses aid in clustering segment features appropriately, but relying solely on them requires iterative
post-processing clustering algorithms such as DBSCAN \cite{schubert2017dbscan}, mean-shift clustering \cite{derpanis2005mean} etc.
In this work, we propose a much simpler the clustering algorithm for projecting features to labels (Section~\ref{sec:ccl}),
by using penalties on pairwise distances in the segment feature metric space.
To this end, we propose a loss function that focuses on fusable
and separable edges independently (Equations ~\eqref{eqn:seg_base}, ~\eqref{eqn:seg_sep} and ~\eqref{eqn:seg_fuse}).
We reuse the same threshold $\Delta_V$ here to place constraints along the edges of the graph.

\begin{equation}
  \label{eqn:seg_base}
  \mathcal{L}_\text{segment} = w_\text{fuse} \mathcal{L}_\text{fuse} + w_\text{sep} \mathcal{L}_\text{sep}
\end{equation}

\begin{equation}
  \label{eqn:seg_sep}
  \mathcal{L}_\text{sep} = \frac{1}{|\mathcal{E}_\text{sep}|} \sum_{e_{ij} \in \mathcal{E}_\text{sep}} H( \Delta_V - d(\bar{\mathbf{F}_i}, \bar{\mathbf{F}_j}))
\end{equation}

\begin{equation}
  \label{eqn:seg_fuse}
  \mathcal{L}_\text{fuse} = \frac{1}{|\mathcal{E}_\text{fuse}|} \sum_{e_{ij} \in \mathcal{E}_\text{fuse}} H( d(\bar{\mathbf{F}_i}, \bar{\mathbf{F}_j}) - \Delta_V)
\end{equation}

where $\mathcal{E}_\text{sep}$ and $\mathcal{E}_\text{fuse}$ denote the set of edges which should be kept separate and fused
respectively, decided using ground truth instance information. %

Since the number of objects in a scene are finite, the number of edges between dissimilar objects are much lower
than the edges within the same object.
This influences the relative weighing of $w_\text{fuse}$ and $w_\text{sep}$, which we set to be $1$ and $0.01$ respectively, to counter this imbalance.

\subsection{Connected Component Labelling (CCL)}
\label{sec:ccl}
To convert the segment features $\bar{\mathbf{F}}$ to fusion decisions, we employ a connected component labelling algorithm over the
feature space that $\bar{\mathbf{F}}$ resides in. Pairwise-distances are computed $d(\bar{\mathbf{F}_i}, \bar{\mathbf{F}_j})\ \forall\ e_{ij} \in \mathcal{E}$ and
thresholded using $\Delta_V$, depicted through a piecewise function as in Equation~\ref{eqn:thresh}. Thus, each positive entry of
$\mathbf{B}_{ij}$ indicates a fusing decision between segments $i$ and $j$.
\begin{equation}
  \label{eqn:thresh}
  \mathbf{B}_{ij} =
    \begin{cases}
      1 & d(\bar{\mathbf{F}_i}, \bar{\mathbf{F}_j}) < \Delta_V\ \forall\ e_{ij} \in \mathcal{E} \\
      0 & \text{otherwise} \\
    \end{cases}
\end{equation}
For every pair of indices for which $\mathbf{B}_{ij} = 1$, we use an efficient disjoint-set forest to implement Union-Find to record the
connectivity information \cite{cormen2009introduction}. For every connected component composed of a set of segments,
we compute the mean semantic probability vector of the component and identify the most probable semantic label for the entire component,
thus forming the final set of predictions $\mathbf{S}_{SF}$.
This procedure enables us to (i) identify and assign a single label to an entire object, and (ii) allow the \sf
block to override erroneous point-wise semantic predictions for parts of the object by using consensus information
across larger parts of the object.

\begin{table*}[t!]
    \caption{Performance impact (mIoU) of \sf on state-of-the-art semantic segmentation backbones on the ScanNet.}
    \tablenegspace
  \begin{center}
    \resizebox{\textwidth}{!}{%
    \begin{tabular}{ccccccccccccccccccccccc}
      \toprule
        Model & \rot{90}{Set} & \rot{90}{wall} & \rot{90}{floor} & \rot{90}{cabinet} & \rot{90}{bed} & \rot{90}{chair} & \rot{90}{sofa} & \rot{90}{table} & \rot{90}{door} & \rot{90}{window} & \rot{90}{bookshelf} & \rot{90}{picture} & \rot{90}{counter} & \rot{90}{desk} & \rot{90}{curtain} & \rot{90}{refridgerator} & \rot{90}{showercurtain} & \rot{90}{toilet} & \rot{90}{sink} & \rot{90}{bathtub} & \rot{90}{otherfurniture} & \rot{90}{\textbf{mean}} \\
        \midrule
        SparseConvNet & val & 82.4 & 94.8 & 58.0 & 77.2 & 88.5 & 78.4 & 68.1 & 58.3 & 58.7 & 70.1 & 29.4 & 61.9 & 56.7 & 62.3 & 46.6 & 61.5 & 91.2 & 63.7 & 84.0 & 50.3 & 67.1\\
        SparseConvNet + SF & val & 85.3 & 97.1 & 61.9 & 79.5 & 91.6 & 82.1 & 72.3 & 61.4 & 61.0 & 71.6 & 36.1 & 71.0 & 61.7 & 65.8 & 47.1 & 69.9 & 95.5 & 73.8 & 92.3 & 54.0 & \textbf{71.5}\\
        \midrule
        PointConv & val & 74.1 & 94.7 & 46.9 & 70.0 & 83.0 & 70.0 & 64.9 & 32.1 & 46.7 & 68.4 & 11.7 & 56.6 & 52.4 & 58.2 & 36.6 & 46.8 & 83.2 & 58.0 & 77.3 & 34.6 & 58.3 \\
        PointConv + SF & val & 78.6 & 97.3 & 50.7 & 78.0 & 87.6 & 76.2 & 68.0 & 36.0 & 49.3 & 75.1 & 13.3 & 64.2 & 59.9 & 62.9 & 38.1 & 54.1 & 89.9 & 63.6 & 90.5 & 38.2 & \textbf{63.6}\\
        \midrule
        MinkNet42 & val & 84.3 & 95.1 & 63.3 & 78.9 & 91.5 & 87.7 & 74.4 & 60.2 & 65.0 & 80.1 & 24.9 & 65.1 & 65.8 & 78.1 & 55.4 & 69.9 & 92.2 & 69.1 & 86.4 & 61.0 & 72.4\\
        MinkNet42 + SF & val & 86.7 & 97.3 & 65.9 & 80.0 & 93.9 & 90.0 & 77.0 & 63.6 & 66.4 & 82.9 & 26.5 & 68.9 & 68.7 & 80.7 & 55.7 & 72.8 & 95.4 & 76.0 & 91.7 & 64.5 & \textbf{75.2} \\
        \midrule
       MinkNet42 & test & 83.9 & 95.4 & 71.1 & 82.4 & 84.5 & 76.1 & 66 & 59 & 66.8 & 80 & 19.7 & 50.5 & 59.4 & 82.1 & 19.7 & 89.1 & 88.7 & 72.8 & 91.2 & 54.5 & 72.4 \\
       MinkNet42 + SF & test & 85.6 & 97.5 & 72.5 & 83.2 & 86.3 & 77.4 & 67.4 & 63 & 67.7 & 81.2 & 22 & 54.2 & 59.3 & 82.9 & 22 & 93.4 & 92.6 & 77.6 & 97.4 & 56.1 & \textbf{74.7} \\
       \bottomrule
  \end{tabular}%
}
    \label{tab:other_sem_seg_scan}
  \end{center}
\end{table*}
\begin{figure*}[t]
  \centering
  \includegraphics[width=\textwidth]{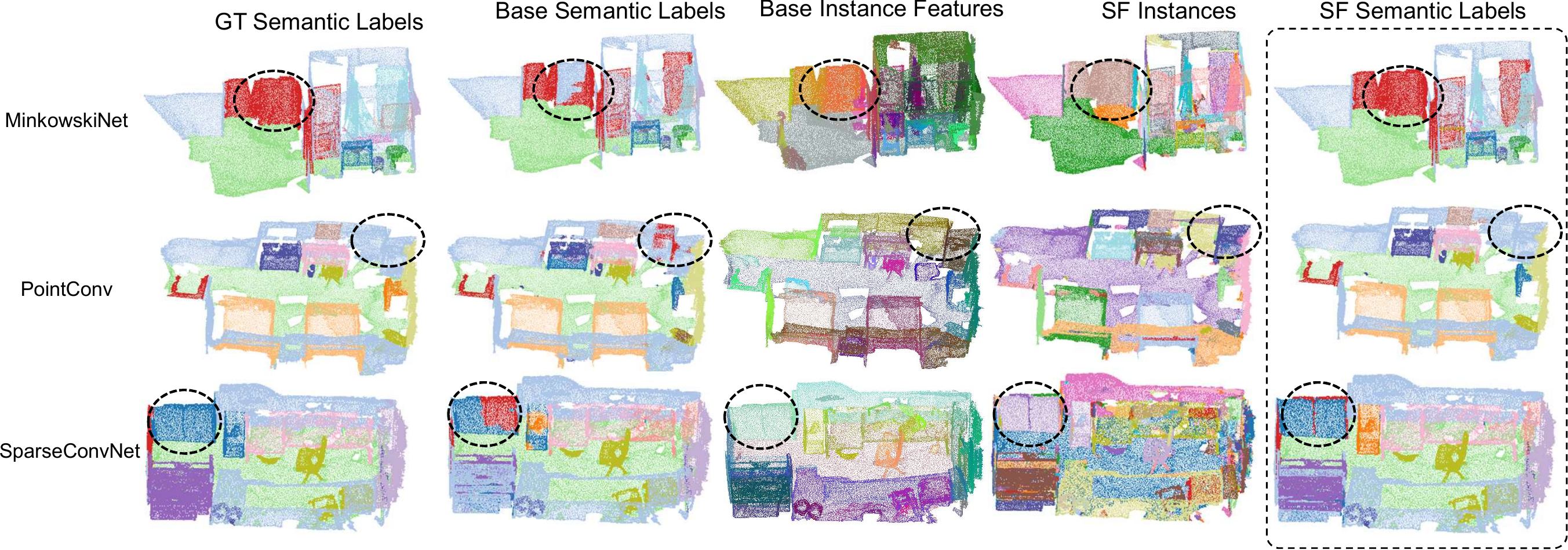}
  \caption{Qualitative results of \sf on some sample point clouds of the ScanNet validation set using the MinkowskiNet-42 \cite{choy20194d}, PointConv \cite{wu2019pointconv} and SparseConvNet \cite{graham20183d} semantic backbones. The highlighted regions indicate mispredictions in the base semantic model ($\mathbf{S}$). Note how \sf builds consensus using the instance features to correct the semantic predictions in these areas (without degrading the performance in the other regions).}
  \label{fig:seg_fusion_egs}
\end{figure*}
\section{Evaluation}
\subsection{Training Setup}

We apply and evaluate the proposed \sf method with multiple state-of-the-art backbones of semantic segmentation across ScanNet and S3DIS datasets using the mean Intersection-over-Union (mIoU) metric.
To assess qualitative improvements with Segmet-Fusion, we also compare results with and without \sf (and show how it helps in regions with part misclassification).
\subsection{ScanNet Dataset}
We pick three state-of-the-art semantic segmentation backbones -- SparseConvNet ~\cite{graham20183d},
PointConv ~\cite{wu2019pointconv} and MinkowskiNet ~\cite{choy20194d}.
We augment each of these semantic backbones with an instance backbone trained with the losses proposed in Occuseg ~\cite{han2020occuseg}.
On the ScanNet validation set, as shown in Table~\ref{tab:other_sem_seg_scan},
we observe significant improvements of $4.4\%$, $5.1\%$ and $2.8\%$ respectively in mIoU scores.
Also, we notice a consistent upswing in IoU scores across all classes. 
Qualitative improvements can be observed in illustrations of specific point clouds
in Figure~\ref{fig:seg_fusion_egs}.
Thus, we observe that the improvements provided by \sf do not depend specifically on the choice of the semantic segmentation backbone used.
On the ScanNet test set, we applied \sf on the MinkowskiNet semantic backbone and obtained the $5^{th}$ position on the leaderboard ($74.7\%$ mIoU), obtaining a significant improvement of $2.3\%$ in mIoU score.

\subsection{S3DIS Dataset}
\begin{figure*}
\begin{minipage}{\textwidth}
\begin{minipage}[h]{0.6\textwidth}
\captionsetup{width=\textwidth}
  \captionof{table}{Impact of \sf on state-of-the-art semantic segmentation backbones on the S3DIS Area 5}
  \tablenegspace
    \resizebox{\textwidth}{!}{%
    \begin{tabular}{ccccccccccccccc}
      \toprule
      Model & \rot{90}{clutter} & \rot{90}{beam} & \rot{90}{board} & \rot{90}{bookcase} & \rot{90}{ceiling} & \rot{90}{chair} & \rot{90}{column} & \rot{90}{door} & \rot{90}{floor} & \rot{90}{sofa} & \rot{90}{table} & \rot{90}{wall} & \rot{90}{window} & \rot{90}{mean} \\

      \midrule
        MinkNet18 & 51.4 & 0.0 & 66.5 & 67.4 & 92.3 & 87.1 & 34.9 & 68.4 & 95.9 & 59.8 & 76.0 & 81.0 & 49.2 & 63.8 \\
        MinkNet18 + SF & 52.4 & 0.0 & 69.3 & 67.2 & 93.0 & 88.4 & 37.6 & 67.7 & 96.7 & 67.7 & 76.5 & 81.6 & 51.2 & 65.3 \\
      \midrule
        KPConv & 54.3 & 0.0 & 60.9 & 72.5 & 92.3 & 89.7 & 20.2 & 73.2 & 96.8 & 70.1 & 77.9 & 80.1 & 52.0 & 64.6 \\
      KPConv + SF & 55.2 & 0.0 & 61.8 & 73.3 & 92.4 & 89.5 & 20.7 & 75.6 & 96.9 & 70.0 & 77.9 & 80.7 & 53.4 & 65.2\\
      \midrule
      ASIS & 41.5 & 0.0 & 52.5 & 0.0 & 89.6 & 1.4 & 7.8 & 17.2 & 96.5 & 0.0 & 1.4 & 73.1 & 39.9 & 32.4 \\
      ASIS + SF & 43.2 & 0.0 & 61.6 & 0.0 & 89.7 & 0.0 & 2.9 & 16.9 & 96.6 & 0.0 & 0.0 & 74.5 & 43.4 & 33.2 \\
      \bottomrule
  \end{tabular}%
}
    \label{tab:other_sem_seg_s3dis}
\end{minipage}
\begin{minipage}[h]{0.4\textwidth}
  \centering
  \captionsetup{width=0.85\textwidth}
  \includegraphics[width=\textwidth]{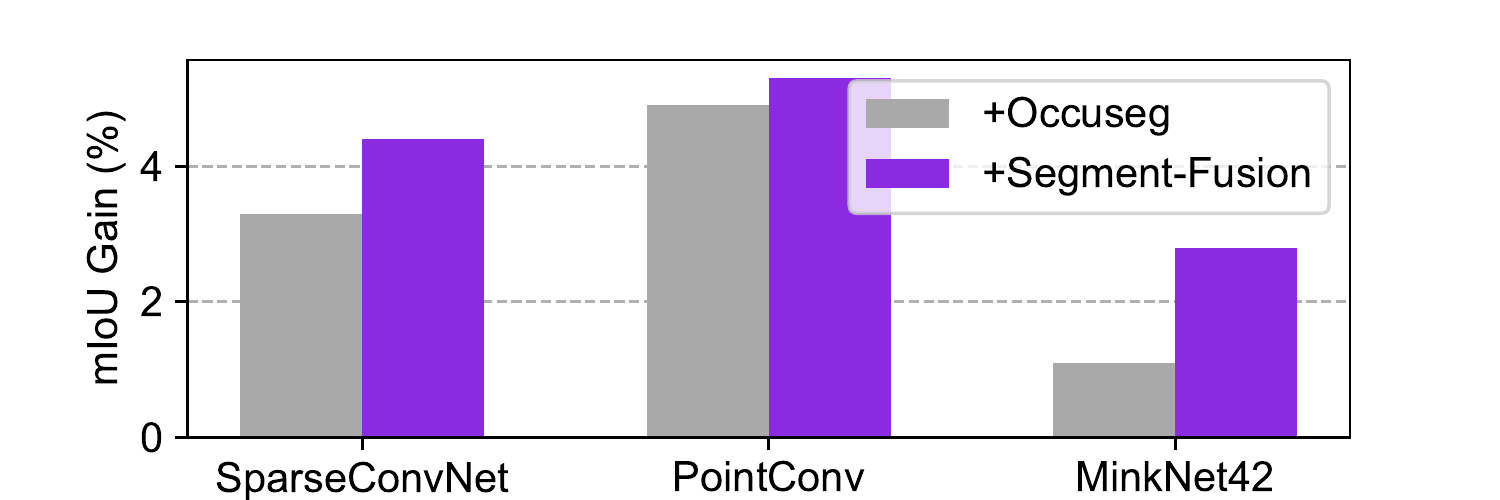}
  \captionof{figure}{Comparison between the semantic segmentation performance of Occuseg's clustering algorithm and \sf's learnable fusion algorithm.}
  \label{fig:occuseg_cmp}
\end{minipage}
\end{minipage}
\end{figure*}

To demonstrate the effectiveness of our approach on datasets that do not have mesh information, we evaluate \sf on the Area 5 set from S3DIS dataset.
We choose three semantic backbone networks: MinkNet18 \cite{choy20194d}, KPConv \cite{thomas2019kpconv} and ASIS \cite{wang2019associatively} to show improvements with \sf.
Table~\ref{tab:other_sem_seg_s3dis} presents semantic segmentation results on the S3DIS Area 5 set on a variety of semantic
backbones. We augment each of these semantic backbones with instance features
from ASIS \cite{wang2019associatively}. The results show that we obtain $1.5\%$, $0.6\%$ and $0.8\%$ improvement
in mIoU on using \sf over MinkNet18 \cite{choy20194d}, KPConv \cite{thomas2019kpconv} and ASIS \cite{wang2019associatively} respectively.
\subsection{Comparisons with Occuseg clustering}
We also assess \sf with heuristic driven graph clustering approach as employed in OccuSeg \cite{han2020occuseg}.
For this we use semantic and instance backbones similar to OccuSeg, and compare the improvement in semantic segmentation performance
obtained through the iterative clustering algorithm they propose, with that of \sf. %
Figure~\ref{fig:occuseg_cmp} shows that \sf
outperforms the heuristic driven iterative clustering algorithm across semantic backbones; lending the conclusion that a learnable algorithm
like \sf may be more scalable and generic to apply on generic semantic and instance backbones as compared to a hand-crafted algorithm.

\subsection{Ablation Studies}

\subsubsection{Impact of learning beyond J-GS}
To better understand the effectiveness of our approach, we study the individual contributions of the components
  of our system. Table~\ref{tab:sf_sens} compares the mIoU of the 3 semantic segmentation backbones employing
  (i) only graph segmentation based voting (J-GSV) without any fusion network, (ii) J-GS along with \sf (SF), but trained only with instance losses ($\mathcal{L}_\text{instance}$),
  and (iii) SF trained with both segment and instance losses ($\mathcal{L}_\text{segment} + \mathcal{L}_\text{instance}$). In J-GSV, we obtain semantic labels by performing
  majority voting within the segment; while in SF, semantic labels are computed as described in Section~\ref{sec:ccl}.
  It is noteworthy that employing only $\mathcal{L}_\text{instance}$ degrades performance, essentially because we use a simplistic CCL
  algorithm to compute labels from the features; using only instance loss results in over-fusion of segments, leading to poorer performance.
  We observe a significant performance improvement on adding segment losses to constrain individual edges.
  Thus, noting steady improvements in mIoU on account of J-GS and SF, we conclude that (i) a hierarchical approach consisting of
  abstracting pointwise features to segment-wise features and (ii) learnable methods to fuse these segments
  are both required and effective. 
\begin{table}[h]
  \caption{Impact of individual components of \sf on overall mIoU (ScanNet validation set) ($\mathcal{L}_\text{ins} = \mathcal{L}_\text{instance}$, $\mathcal{L}_\text{seg} = \mathcal{L}_\text{segment}$)}
  \tablenegspace
  \begin{center}
    \resizebox{1.0\columnwidth}{!}{%
      \tiny
      \begin{tabular}{@{}ccccc@{}}
        \toprule
        \shortstack{\textbf{Model}\\ \phantom{J}} & \shortstack{\textbf{Base}\\ \phantom{J}} & \shortstack{\textbf{Base +}\\ \textbf{J-GSV}} & \shortstack{\textbf{Base +}\\ \textbf{SF}($\mathcal{L}_\text{ins}$)} & \shortstack{\textbf{Base +}\\ \textbf{SF}($\mathcal{L}_\text{ins}\!\!+\!\!\mathcal{L}_\text{seg}$)} \\
        \midrule
        SparseConvNet & 67.1 & 70.4 & 69.7 & 71.5 \\
        PointConv & 58.3 & 61.5 & 61.4 & 63.6 \\
        MinkNet42 & 72.4 & 74.6 & 73.11 & 75.2 \\
        \bottomrule
  \end{tabular}%
  }
  \label{tab:sf_sens}
  \end{center}
\end{table}
\vspace{-10mm}

\subsubsection{Does J-GS result in over-fusion?}
\begin{figure}[h]
  \centering
  \includegraphics[width=1.0\columnwidth]{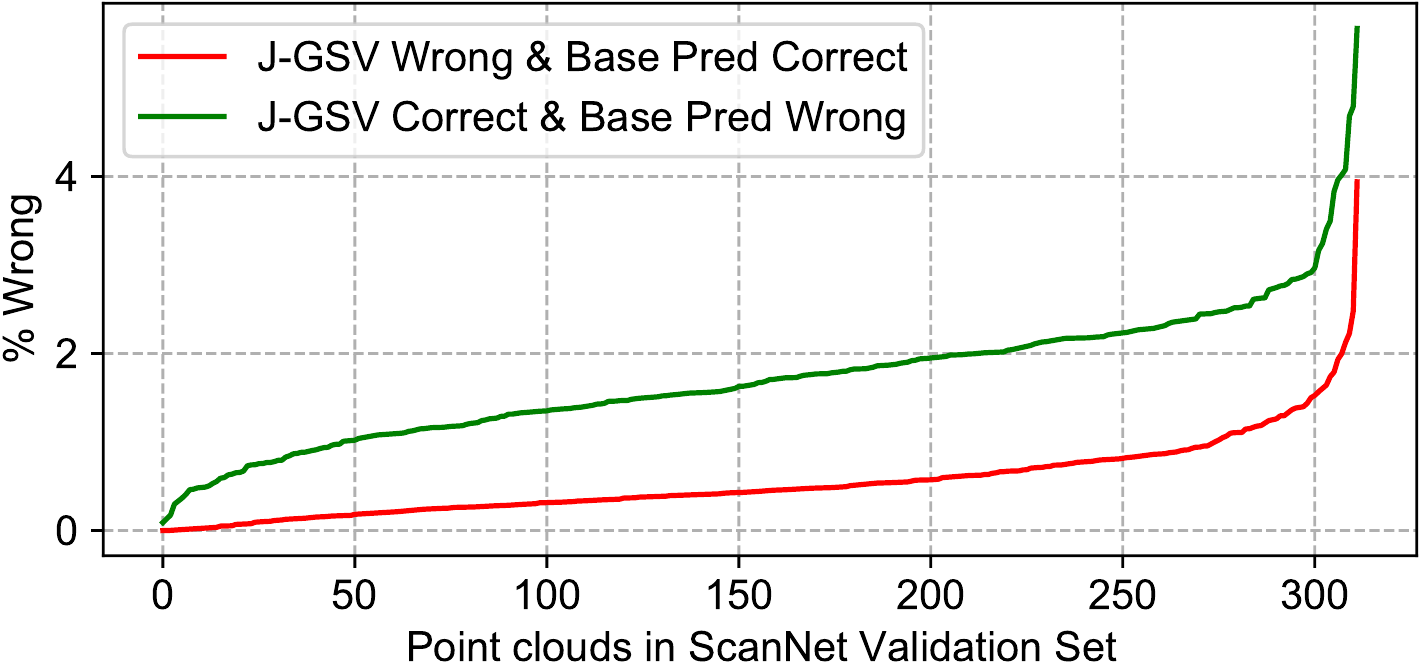}
  \caption{Top: Fraction of points in a point cloud where J-GSV is incorrect and the base semantic predictor is correct.
           Bottom: Fraction points in a point cloud where J-GSV is correct and the base semantic predictor is incorrect.}
  \label{fig:scannet_gsv_val_analysis}
\end{figure}

  While graph segmentation aids in abstracting the problem to a higher level spatially, it runs the risk of over-fusing
  points belonging to two different classes into one segment. In this section, we evaluate
  the impact of such occurrences on overall performance using the proposed J-GSV score.
  For each point cloud, we analyse the base semantic predictions $\mathbf{S}$ with the predictions after J-GSV $\mathbf{S}_{maj}$(without \sf Network)
  and count the number of points for which they disagree with each other.
  We define overfusion to occur when the J-GSV predictions are incorrect while the base predictor is correct.
  In Figure~\ref{fig:scannet_gsv_val_analysis} we observe that J-GS based voting helps in an improvement of atleast $2\%$ in $\sim32\%$ of the point clouds in the validation set
  of the ScanNet dataset, while being $2\%$ incorrect in $\sim1\%$ of the point clouds. This indicates that J-GSV is consistently beneficial
  across point clouds of the dataset and provides higher improvement than the degradation due to overfusion.

\subsubsection{J-GS on S3DIS}
\begin{figure}
  \centering
  \includegraphics[width=0.5\textwidth]{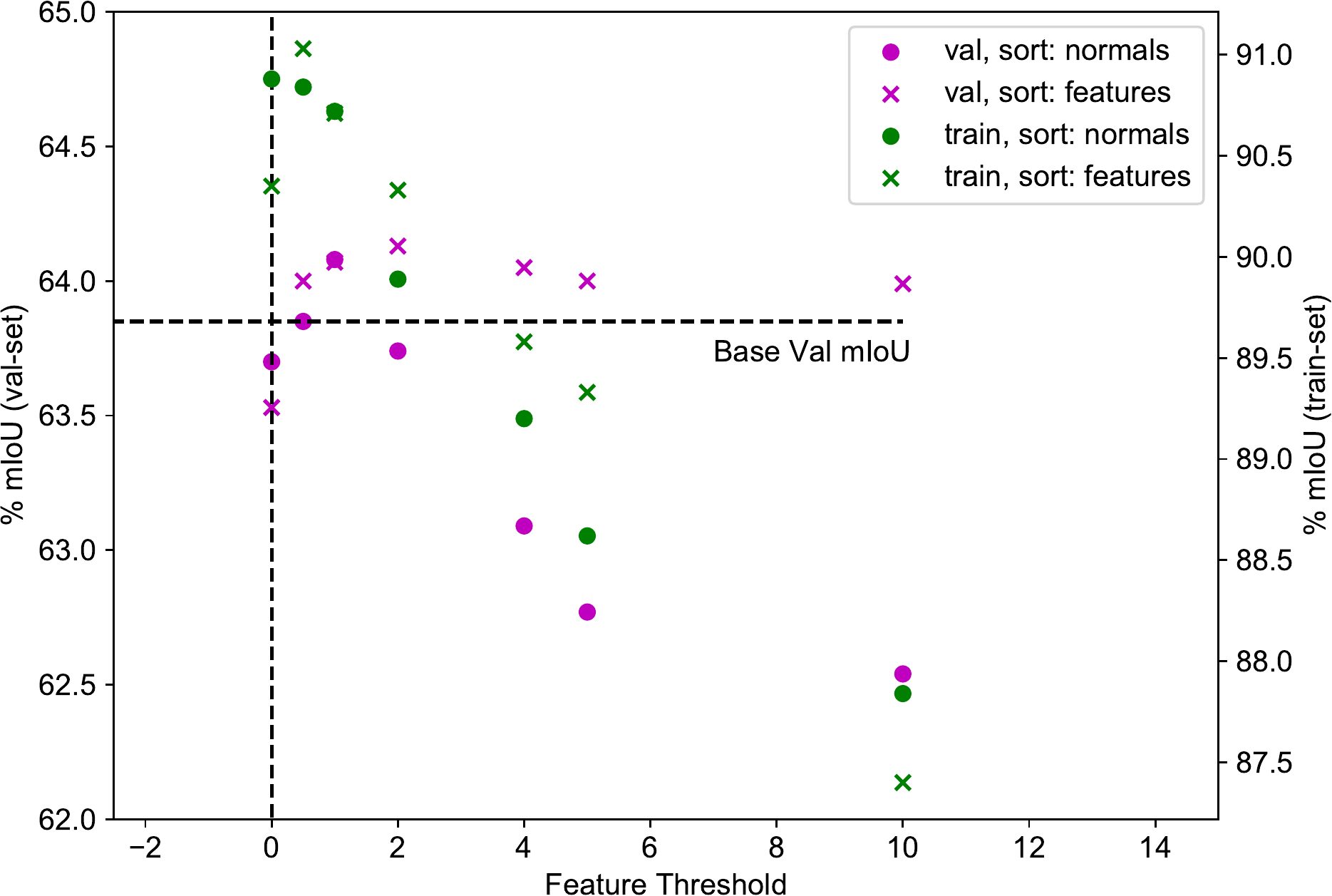}
  \caption{J-GS Voting Score results for the S3DIS validation set across a variety of feature thresholds.
  The threshold used in the normals space $n_{th}$ is $0.01$; results along this axes are omitted here due to lack of sensitivity.}
  \label{fig:s3dis_gs_voting_score_plot}
\end{figure}

We present an ablative study to describe the methodology of selecting the thresholds for  the proposed J-GS method.
Figure~\ref{fig:s3dis_gs_voting_score_plot} illustrates the performance of J-GSV on the training and validation sets of the S3DIS dataset.
As discussed in Section~\ref{sec:gsv_method}, we evaluate the performance of various J-GSV variants on the training set
to select the J-GS parameters and apply them to the validation set. It is observed that on an average, sorting of edge weights along
feature similarity performs better than sorting along similarity along normals.
  Also, relying solely on normals ends up reducing performance as compared to the base validation performance (without J-GSV).
  We observe that $(s:feats,f_{th}:0.5)$ provides optimal performance for the training set,
and use this to apply J-GS on the validation set
and generate segment-wise data for \sf Network. To evaluate how well these parameters generalize from the training
to the validation set, we evaluate the same parameter set on the validation set as well. Though $(s:feats,f_{th}:2)$
  performs better on the validation set, the difference with $f_{th}:0.5$ is low (lower than the gains we see with SF downstream, Table~\ref{tab:other_sem_seg_s3dis}). On the validation set,
the sorting parameter $s$ has a higher sensitivity than the feature-threshold parameter.
The ScanNet dataset has distinct normals, hence we did not observe much sensitivity with the $f_{th}$ threshold.

\section{Limitations and Future Work}
While we have demonstrated the effectiveness of our approach on multiple semantic and instance backbones,
the method still requires supervision.
There is also potential in using deeper and more complex network architectures in the Segment-Fusion network block
to increase improvements.
In the future, we plan to extend this work by using similar strategies to fuse learnings from problems
with shared context, along with focusing on semi/self-supervised learning.

\section{Conclusion}
We presented \sf, a learnable, hierarchical method to fuse segments, aimed at
improving semantic segmentation performance of generic base models. We proposed a voting based algorithm
to select optimal graph segmentation hyperparameters for varying datasets. 
Our learnable method can be flexibly used in conjunction with existing semantic/instance models.
Finally, we demonstrated comprehensive
evaluations to demonstrate the effectiveness of our methods on multiple types of semantic backbones.

{\small
\bibliographystyle{ieee_fullname}
\bibliography{egbib}

\begin{thebibliography}{10}\itemsep=-1pt

\bibitem{armeni2017joint}
Iro Armeni, Sasha Sax, Amir~R Zamir, and Silvio Savarese.
\newblock Joint 2d-3d-semantic data for indoor scene understanding.
\newblock {\em arXiv preprint arXiv:1702.01105}, 2017.

\bibitem{chen2021hierarchical}
Shaoyu Chen, Jiemin Fang, Qian Zhang, Wenyu Liu, and Xinggang Wang.
\newblock Hierarchical aggregation for 3d instance segmentation.
\newblock In {\em Proceedings of the IEEE/CVF International Conference on
  Computer Vision}, pages 15467--15476, 2021.

\bibitem{choy20194d}
Christopher Choy, JunYoung Gwak, and Silvio Savarese.
\newblock 4d spatio-temporal convnets: Minkowski convolutional neural networks.
\newblock In {\em Proceedings of the IEEE/CVF Conference on Computer Vision and
  Pattern Recognition}, pages 3075--3084, 2019.

\bibitem{cormen2009introduction}
Thomas~H Cormen, Charles~E Leiserson, Ronald~L Rivest, and Clifford Stein.
\newblock {\em Introduction to algorithms}.
\newblock MIT press, 2009.

\bibitem{dai2017scannet}
Angela Dai, Angel~X Chang, Manolis Savva, Maciej Halber, Thomas Funkhouser, and
  Matthias Nie{\ss}ner.
\newblock Scannet: Richly-annotated 3d reconstructions of indoor scenes.
\newblock In {\em Proceedings of the IEEE conference on computer vision and
  pattern recognition}, pages 5828--5839, 2017.

\bibitem{derpanis2005mean}
Konstantinos~G Derpanis.
\newblock Mean shift clustering.
\newblock {\em Lecture Notes}, page~32, 2005.

\bibitem{engelmann20203d}
Francis Engelmann, Martin Bokeloh, Alireza Fathi, Bastian Leibe, and Matthias
  Nie{\ss}ner.
\newblock 3d-mpa: Multi-proposal aggregation for 3d semantic instance
  segmentation.
\newblock In {\em Proceedings of the IEEE/CVF conference on computer vision and
  pattern recognition}, pages 9031--9040, 2020.

\bibitem{felzenszwalb2004efficient}
Pedro~F Felzenszwalb and Daniel~P Huttenlocher.
\newblock Efficient graph-based image segmentation.
\newblock {\em International journal of computer vision}, 59(2):167--181, 2004.

\bibitem{gong2021omni}
Jingyu Gong, Jiachen Xu, Xin Tan, Haichuan Song, Yanyun Qu, Yuan Xie, and
  Lizhuang Ma.
\newblock Omni-supervised point cloud segmentation via gradual receptive field
  component reasoning.
\newblock In {\em Proceedings of the IEEE/CVF Conference on Computer Vision and
  Pattern Recognition}, pages 11673--11682, 2021.

\bibitem{graham20183d}
Benjamin Graham, Martin Engelcke, and Laurens Van Der~Maaten.
\newblock 3d semantic segmentation with submanifold sparse convolutional
  networks.
\newblock In {\em Proceedings of the IEEE conference on computer vision and
  pattern recognition}, pages 9224--9232, 2018.

\bibitem{gupta2015indoor}
Saurabh Gupta, Pablo Arbel{\'a}ez, Ross Girshick, and Jitendra Malik.
\newblock Indoor scene understanding with rgb-d images: Bottom-up segmentation,
  object detection and semantic segmentation.
\newblock {\em International Journal of Computer Vision}, 112(2):133--149,
  2015.

\bibitem{han2020occuseg}
Lei Han, Tian Zheng, Lan Xu, and Lu Fang.
\newblock Occuseg: Occupancy-aware 3d instance segmentation.
\newblock In {\em Proceedings of the IEEE/CVF conference on computer vision and
  pattern recognition}, pages 2940--2949, 2020.

\bibitem{hou20193d}
Ji Hou, Angela Dai, and Matthias Nie{\ss}ner.
\newblock 3d-sis: 3d semantic instance segmentation of rgb-d scans.
\newblock In {\em Proceedings of the IEEE/CVF Conference on Computer Vision and
  Pattern Recognition}, pages 4421--4430, 2019.

\bibitem{hu2020class}
Hanzhe Hu, Deyi Ji, Weihao Gan, Shuai Bai, Wei Wu, and Junjie Yan.
\newblock Class-wise dynamic graph convolution for semantic segmentation.
\newblock In {\em Computer Vision--ECCV 2020: 16th European Conference,
  Glasgow, UK, August 23--28, 2020, Proceedings, Part XVII 16}, pages 1--17.
  Springer, 2020.

\bibitem{hu2021bidirectional}
Wenbo Hu, Hengshuang Zhao, Li Jiang, Jiaya Jia, and Tien-Tsin Wong.
\newblock Bidirectional projection network for cross dimension scene
  understanding.
\newblock In {\em Proceedings of the IEEE/CVF Conference on Computer Vision and
  Pattern Recognition}, pages 14373--14382, 2021.

\bibitem{hu2021vmnet}
Zeyu Hu, Xuyang Bai, Jiaxiang Shang, Runze Zhang, Jiayu Dong, Xin Wang,
  Guangyuan Sun, Hongbo Fu, and Chiew-Lan Tai.
\newblock Vmnet: Voxel-mesh network for geodesic-aware 3d semantic
  segmentation.
\newblock In {\em Proceedings of the IEEE/CVF International Conference on
  Computer Vision}, pages 15488--15498, 2021.

\bibitem{hu2020jsenet}
Zeyu Hu, Mingmin Zhen, Xuyang Bai, Hongbo Fu, and Chiew-lan Tai.
\newblock Jsenet: Joint semantic segmentation and edge detection network for 3d
  point clouds.
\newblock In {\em Computer Vision--ECCV 2020: 16th European Conference,
  Glasgow, UK, August 23--28, 2020, Proceedings, Part XX 16}, pages 222--239.
  Springer, 2020.

\bibitem{jiang2020pointgroup}
Li Jiang, Hengshuang Zhao, Shaoshuai Shi, Shu Liu, Chi-Wing Fu, and Jiaya Jia.
\newblock Pointgroup: Dual-set point grouping for 3d instance segmentation.
\newblock In {\em Proceedings of the IEEE/CVF Conference on Computer Vision and
  Pattern Recognition}, pages 4867--4876, 2020.

\bibitem{klasing2008clustering}
Klaas Klasing, Dirk Wollherr, and Martin Buss.
\newblock A clustering method for efficient segmentation of 3d laser data.
\newblock In {\em 2008 IEEE international conference on robotics and
  automation}, pages 4043--4048. IEEE, 2008.

\bibitem{kundu2020virtual}
Abhijit Kundu, Xiaoqi Yin, Alireza Fathi, David Ross, Brian Brewington, Thomas
  Funkhouser, and Caroline Pantofaru.
\newblock Virtual multi-view fusion for 3d semantic segmentation.
\newblock In {\em European Conference on Computer Vision}, pages 518--535.
  Springer, 2020.

\bibitem{liang2021instance}
Zhihao Liang, Zhihao Li, Songcen Xu, Mingkui Tan, and Kui Jia.
\newblock Instance segmentation in 3d scenes using semantic superpoint tree
  networks.
\newblock In {\em Proceedings of the IEEE/CVF International Conference on
  Computer Vision}, pages 2783--2792, 2021.

\bibitem{nekrasov2021mix3d}
Alexey Nekrasov, Jonas Schult, Or Litany, Bastian Leibe, and Francis Engelmann.
\newblock Mix3d: Out-of-context data augmentation for 3d scenes.
\newblock {\em arXiv preprint arXiv:2110.02210}, 2021.

\bibitem{pham2019jsis3d}
Quang-Hieu Pham, Thanh Nguyen, Binh-Son Hua, Gemma Roig, and Sai-Kit Yeung.
\newblock Jsis3d: Joint semantic-instance segmentation of 3d point clouds with
  multi-task pointwise networks and multi-value conditional random fields.
\newblock In {\em Proceedings of the IEEE/CVF Conference on Computer Vision and
  Pattern Recognition}, pages 8827--8836, 2019.

\bibitem{qi2017pointnet}
Charles~R Qi, Hao Su, Kaichun Mo, and Leonidas~J Guibas.
\newblock Pointnet: Deep learning on point sets for 3d classification and
  segmentation.
\newblock In {\em Proceedings of the IEEE conference on computer vision and
  pattern recognition}, pages 652--660, 2017.

\bibitem{qi2017pointnet++}
Charles~R Qi, Li Yi, Hao Su, and Leonidas~J Guibas.
\newblock Pointnet++: Deep hierarchical feature learning on point sets in a
  metric space.
\newblock {\em arXiv preprint arXiv:1706.02413}, 2017.

\bibitem{schubert2017dbscan}
Erich Schubert, J{\"o}rg Sander, Martin Ester, Hans~Peter Kriegel, and Xiaowei
  Xu.
\newblock Dbscan revisited, revisited: why and how you should (still) use
  dbscan.
\newblock {\em ACM Transactions on Database Systems (TODS)}, 42(3):1--21, 2017.

\bibitem{siam2018comparative}
Mennatullah Siam, Mostafa Gamal, Moemen Abdel-Razek, Senthil Yogamani, Martin
  Jagersand, and Hong Zhang.
\newblock A comparative study of real-time semantic segmentation for autonomous
  driving.
\newblock In {\em Proceedings of the IEEE conference on computer vision and
  pattern recognition workshops}, pages 587--597, 2018.

\bibitem{strom2010graph}
Johannes Strom, Andrew Richardson, and Edwin Olson.
\newblock Graph-based segmentation for colored 3d laser point clouds.
\newblock In {\em 2010 IEEE/RSJ international conference on intelligent robots
  and systems}, pages 2131--2136. IEEE, 2010.

\bibitem{thomas2019kpconv}
Hugues Thomas, Charles~R Qi, Jean-Emmanuel Deschaud, Beatriz Marcotegui,
  Fran{\c{c}}ois Goulette, and Leonidas~J Guibas.
\newblock Kpconv: Flexible and deformable convolution for point clouds.
\newblock In {\em Proceedings of the IEEE/CVF International Conference on
  Computer Vision}, pages 6411--6420, 2019.

\bibitem{valada2017adapnet}
Abhinav Valada, Johan Vertens, Ankit Dhall, and Wolfram Burgard.
\newblock Adapnet: Adaptive semantic segmentation in adverse environmental
  conditions.
\newblock In {\em 2017 IEEE International Conference on Robotics and Automation
  (ICRA)}, pages 4644--4651. IEEE, 2017.

\bibitem{vaswani2017attention}
Ashish Vaswani, Noam Shazeer, Niki Parmar, Jakob Uszkoreit, Llion Jones,
  Aidan~N Gomez, {\L}ukasz Kaiser, and Illia Polosukhin.
\newblock Attention is all you need.
\newblock In {\em Advances in neural information processing systems}, pages
  5998--6008, 2017.

\bibitem{wang2019graph}
Lei Wang, Yuchun Huang, Yaolin Hou, Shenman Zhang, and Jie Shan.
\newblock Graph attention convolution for point cloud semantic segmentation.
\newblock In {\em Proceedings of the IEEE/CVF Conference on Computer Vision and
  Pattern Recognition}, pages 10296--10305, 2019.

\bibitem{wang2017cnn}
Peng-Shuai Wang, Yang Liu, Yu-Xiao Guo, Chun-Yu Sun, and Xin Tong.
\newblock O-cnn: Octree-based convolutional neural networks for 3d shape
  analysis.
\newblock {\em ACM Transactions On Graphics (TOG)}, 36(4):1--11, 2017.

\bibitem{wang2018sgpn}
Weiyue Wang, Ronald Yu, Qiangui Huang, and Ulrich Neumann.
\newblock Sgpn: Similarity group proposal network for 3d point cloud instance
  segmentation.
\newblock In {\em Proceedings of the IEEE conference on computer vision and
  pattern recognition}, pages 2569--2578, 2018.

\bibitem{wang2019associatively}
Xinlong Wang, Shu Liu, Xiaoyong Shen, Chunhua Shen, and Jiaya Jia.
\newblock Associatively segmenting instances and semantics in point clouds.
\newblock In {\em Proceedings of the IEEE/CVF Conference on Computer Vision and
  Pattern Recognition}, pages 4096--4105, 2019.

\bibitem{wu2019pointconv}
Wenxuan Wu, Zhongang Qi, and Li Fuxin.
\newblock Pointconv: Deep convolutional networks on 3d point clouds.
\newblock In {\em Proceedings of the IEEE/CVF Conference on Computer Vision and
  Pattern Recognition}, pages 9621--9630, 2019.

\bibitem{yu2018ds}
Chao Yu, Zuxin Liu, Xin-Jun Liu, Fugui Xie, Yi Yang, Qi Wei, and Qiao Fei.
\newblock Ds-slam: A semantic visual slam towards dynamic environments.
\newblock In {\em 2018 IEEE/RSJ International Conference on Intelligent Robots
  and Systems (IROS)}, pages 1168--1174. IEEE, 2018.

\bibitem{zhang2019dual}
Li Zhang, Xiangtai Li, Anurag Arnab, Kuiyuan Yang, Yunhai Tong, and Philip~HS
  Torr.
\newblock Dual graph convolutional network for semantic segmentation.
\newblock {\em arXiv preprint arXiv:1909.06121}, 2019.

\end{thebibliography}
}

\end{document}